\pdfoutput=1

\documentclass[11pt]{article}

\usepackage{acl}

\usepackage{times}
\usepackage{latexsym}

\usepackage[T1]{fontenc}

\usepackage[utf8]{inputenc}

\usepackage{microtype}

\usepackage{url}
\usepackage{multirow}
\usepackage{makecell}
\usepackage{tabu}
\usepackage{amsfonts}

\usepackage{mathtools}
\usepackage{threeparttable}
\usepackage{bm}

\usepackage{setspace}
\usepackage{enumitem}

\usepackage{times}
\usepackage{latexsym}
\usepackage{multicol}
\usepackage{booktabs}

\usepackage{todonotes}
\usepackage{comment}

\title{Measure and Improve Robustness in NLP Models: A Survey}

\author{Xuezhi Wang \\
  Google Research \\
  {\tt xuezhiw@google.com} \\\And
  Haohan Wang\\
  Carnegie Mellon University\\
  {\tt haohanw@cs.cmu.edu} \\ \And
  Diyi Yang \\
  \ Georgia Institute of Technology\\
  {\tt \ dyang888@gatech.edu} \\}

\begin{document}
\maketitle
\begin{abstract}
As NLP models achieved state-of-the-art performances over benchmarks and gained wide applications, it has been increasingly important to ensure the safe deployment of these models in the real world, e.g., making sure the models are robust against unseen or challenging scenarios.
Despite robustness being an increasingly studied topic, it has been separately explored in applications like vision and NLP, with various definitions, evaluation and mitigation strategies in multiple lines of research.
In this paper, we aim to provide a unifying survey of how to define, measure and improve robustness in NLP.
We first connect multiple definitions of robustness, then unify various lines of work on identifying robustness failures and evaluating models' robustness.
Correspondingly, we present mitigation strategies that are data-driven, model-driven, and inductive-prior-based, with a more systematic view of how to effectively improve robustness in NLP models.
Finally, we conclude by outlining open challenges and future directions to motivate further research in this area.
\end{abstract}

\section{Introduction}
NLP models, especially with the recent advances of large pre-trained language
models have achieved great progress and gained wide applications in the real world. 
Despite the performance gains, NLP models are still fragile and brittle to out-of-domain data \cite{ood_robustness,wang-etal-2019-adversarial-domain}, adversarial attacks \cite{mccoy-etal-2019-right,jia2017adversarial,textfooler}, or small perturbation to the input \cite{ebrahimi-etal-2018-hotflip,belinkov2018synthetic}.
Those failures could hinder the safe deployment of these models in the real world, and impact NLP models' trustworthiness to users.
As a result, an increasing line of work has been conducted to understand robustness issues in the language technologies communities. 
Still, diverse sets of research across multiple dimensions and numerous levels of depth exist and are scattered across various communities; for instance, using a variety of definitions on a wide range of very different NLP tasks. 
In this work, we provide a unifying overview 
of what is robustness in NLP, how to identify robustness failures and evaluate model's robustness, and systematic ways to improve robustness, as well as a conceptual schema categorizing ongoing research directions. 
We identify gaps between the to-date robustness work, the technical opportunities, and discuss possible paths forward. 

\section{Definitions of Robustness in NLP}

Robustness, despite its specific definitions in various lines of research, can typically be unified as follows: 
denote the input as $x$, and its associated gold label for the main task as $y$, assume a model $f$ is trained on $(x, y)\sim \mathcal{D}$ and its prediction over $x$ as $f(x)$; now given test data $(x', y')\sim \mathcal{D}'\neq \mathcal{D}$, we can measure a model's robustness by its performance on $\mathcal{D}'$, e.g., using the model's robust accuracy \cite{tsipras2018robustness,NEURIPS2020_61d77652}, defined as $\mathbb{E}_{(x',y')\sim D'}[f(x')=y']$.
Existing literature on robustness in NLP can be roughly categorized by how $\mathcal{D}'$ is constructed: by synthetically perturbing the input (Section~\ref{sec:adversarial_attack}), or $\mathcal{D}'$ is naturally occurring with a distribution shift (Section~\ref{sec:distribution_shift}).

The above definition works for a range of NLP tasks like text classification and sequence labeling where $y$ is defined over a fixed set of discrete labels. 
For tasks like text generation, robustness is less well defined and can manifest as positional bias \cite{jung-etal-2019-earlier,kryscinski-etal-2019-neural}, or hallucination \cite{maynez-etal-2020-faithfulness,parikh-etal-2020-totto,zhou-etal-2021-detecting}. 
One major challenge here is a lack of robust metrics in evaluating the quality of the generated text \cite{sellam-etal-2020-bleurt,BERTScore}, i.e., we need a reliable metric to determine the relationship between $f(x')$ and $y'$ when both are open-ended texts.

\subsection{Robustness against Adversarial Attacks}
\label{sec:adversarial_attack}
In one line of research, $\mathcal{D}'$ is constructed by perturbations around input $x$ to form $x'$ ($x'$ typically being defined within some proximity of $x$).
This topic has been widely explored in computer vision under the concept of adversarial robustness, which measures models' performances against carefully crafted noises 
generated deliberately to deceive the model to predict wrongly, pioneered by \cite{szegedy2013intriguing,goodfellow2015explaining}, and later extended to NLP, such as \cite{ebrahimi-etal-2018-hotflip,alzantot2018generating,li2018textbugger,feng-etal-2018-pathologies,kuleshov2018adversarial,jia2019certified,zang-etal-2020-word,pruthi2019combating,wang2019natural,garg2020bae,tan-etal-2020-morphin,tan-etal-2020-mind,schwinn2021exploring,li-etal-2021-contextualized,bad_character} and multilingual adversaries \cite{yang-etal-2019-paws,tan-joty-2021-code}. 
The generation of adversarial examples primarily builds upon the observation that we can generate samples that are meaningful to humans (e.g., by perturbing the samples with changes that are imperceptible to humans) while altering the prediction of the models for this sample. 
In this regard, human's remarkable ability in understanding a large set of synonyms \cite{li-etal-2020-bert-attack} or interesting characteristics in ignoring the exact order of letters \cite{wang2020word} are often opportunities to create adversarial examples. 
A related line of work such as data-poisoning \cite{wallace-etal-2021-concealed} and weight-poisoning \cite{kurita-etal-2020-weight} exposes NLP models' vulnerability against attacks during the training process. One can refer to more comprehensive reviews and broader discussions on this topic in \citet{zhang2020adversarial} and \citet{morris2020textattack}.

\noindent
\textbf{Assumptions around Label-preserving and Semantic-preserving}
Most existing work in vision makes a relatively simplified assumption that the gold label of $x'$ remains unchanged under a bounded perturbation over $x$, i.e., $y'=y$, and a model's robust behaviour should be $f(x')=y$ \cite{szegedy2013intriguing,goodfellow2015explaining}.
A similar line of work in NLP follows the same label-preserving assumption with small text perturbations like token and character swapping \cite{alzantot2018generating,textfooler,ren-etal-2019-generating,ebrahimi-etal-2018-hotflip}, paraphrasing \cite{iyyer-etal-2018-adversarial,gan-ng-2019-improving}, semantically equivalent adversarial rules \cite{ribeiro-etal-2018-semantically}, and adding distractors \cite{jia2017adversarial}.
However, this label-preserving assumption might not always hold, e.g., \citet{wang2021adversarial} studied several existing text perturbation techniques and found that a significant portion of perturbed examples are \textbf{not} label-preserving (despite their label-preserving assumptions), or the resulting labels have a high disagreement among human raters (i.e., can even fool humans). 
\citet{morris-etal-2020-reevaluating} also call for more attention to the \textit{validity} of perturbed examples for a more accurate robustness evaluation.

Another line of work aims to perturb the input $x$ to $x'$ in small but meaningful ways that explicitly \textit{change} the gold label, i.e., $y'\neq y$, under which case the robust behaviour of a model should be $f(x') = y'$ and $f(x')\neq y$ \cite{gardner-etal-2020-evaluating, kaushik2019learning,DBLP:conf/aaai/SchlegelNB21}.
We believe these two lines of work are complementary to each other, and both should be explored in future research to measure models' robustness more comprehensively.

One alternative notion is whether the perturbation from $x$ to $x'$ is ``semantic-preseving'' \cite{alzantot2018generating,textfooler,ren-etal-2019-generating} or ``semantic-modifying'' \cite{shi-huang-2020-robustness, jia2017adversarial}. Note this is slightly different from the above label-preserving assumptions, as it is defined over the perturbations on $(x, x')$ rather than making an assumption on $(y, y')$, e.g.,  semantic-modifying perturbations can be either label-preserving \cite{jia2017adversarial,shi-huang-2020-robustness} or label-changing \cite{gardner-etal-2020-evaluating,kaushik2019learning}.

\subsection{Robustness under Distribution Shift}
\label{sec:distribution_shift}
Another line of research focuses on $(x', y')$ drawn from a different distribution that is naturally-occurring \cite{hendrycks2021nae}, where robustness can be defined around model's performance under distribution shift.
Different from work on domain adaptation \cite{patel_vda,wilson_uda} and transfer learning \cite{pan_transfer_learning_survey},
existing definitions of robustness are closer to the concept of domain generalization \cite{pmlr-v28-muandet13,gulrajani2021in}, or out-of-distribution generalization to unforeseen distribution shifts \cite{ood_robustness}, where the test data (either labeled or unlabeled) is assumed not available during training, i.e., generalization without adaptation.
In the context of NLP, robustness to natural distribution shifts can also mean models' performance should not degrade due to the differences in grammar errors, dialects, speakers, languages \cite{oral_language,blodgett-etal-2016-demographic,demszky-etal-2021-learning}, or newly collected datasets for the same task but in different domains \cite{pmlr-v119-miller20a}.
Another closely connected line of research is fairness, which has been studied in various NLP applications, see \cite{sun-etal-2019-mitigating} for a more in-depth survey in this area.
For example, gendered stereotypes or biases have been observed in NLP tasks including co-reference resolution \cite{zhao-etal-2018-gender, rudinger-etal-2017-social}, occupation classification \cite{bias_in_bios}, and neural machine translation \cite{prates2019, DBLP:journals/corr/abs-1901-03116}.

\subsection{Connections and A Common Theme}
The above two categories of robustness can be unified under the same framework, i.e., whether $\mathcal{D}'$ represents a \textit{synthetic} distribution shift (via adversarial attacks) or a \textit{natural} distribution shift.
Existing work has shown a model's performance might degrade substantially in both cases, but the \textit{transferability} of the two categories is relatively under-explored.
In the vision domain, \citet{taori2020measuring} investigate models' robustness to natural distribution shift, and show that robustness to synthetic distribution shift might offer little to no robustness improvement under natural distribution shift. 
Some studies show NLP models might not generalize to \textit{unseen} adversarial patterns \cite{huang-etal-2020-counterfactually,jha2020does,joshi2021investigation}, but more work is needed to systematically bridge the gap between NLP models' robustness under natural and synthetic distribution shifts.

To better understand \textit{why} models exhibit a lack of robustness, some existing work attributed this to the fact that models sometimes utilize \textit{spurious} correlations between input features and labels, rather than the \textit{genuine} ones, where \textit{spurious} features are commonly defined as features that do not causally affect a task's label \cite{pmlr-v119-srivastava20a,zhao_spurious_2020}: they correlate with task labels but fail to transfer to more challenging test conditions or out-of-distribution data \cite{Geirhos_2020}.
Some other work defined it as ``prediction rules that work for the majority examples
but do not hold in general'' \cite{tu2020empirical}. 
Such spurious correlations are sometimes referred as dataset bias \cite{clark-etal-2019-dont,he2019unlearn}, annotation artifacts \cite{gururangan2018annotation}, or group shift \cite{oren2019distributionally} in the literature. 
Further, evidence showed that controlling model's learning in spurious features 
will improve model's performances in distribution shifts \cite{WangGLX19,WangHLX19};
also, discussions on the connections between adversarial robustness and learning of spurious features have been raised \cite{ilyas2019adversarial,wang2020high}. 
Theoretical discussions connecting these fields have also been offered by crediting a reason of model's lack of robustness in either distribution shift or adversarial attack to model's learning of spurious features \cite{wang2021learning}. 

Further, in certain applications, model ``robustness'' can also be connected with models' instability \cite{stability_gupta}, or models having poorly-calibrated uncertainty estimation \cite{chuan2017}, where Bayesian methods \cite{NIPS2011_4329,pmlr-v37-blundell15}, dropout-based \cite{gal2016dropout, kingma2015variational} and ensemble-based approaches \cite{balaji_ensemble} have been proposed to improve models' uncertainty estimation.
Recently, \citet{NEURIPS2019_8558cb40} have shown models' uncertainty estimation can degrade significantly under distributional shift, and call for more work to ensure a model ``knows when it doesn't know'' by giving lower uncertainty estimates over out-of-distribution data.
This is another example where models can be less robust under distributional shifts, and again emphasizes the need of building more unified benchmarks to measure a model's performance (e.g., robust accuracy, calibration, stability) under distribution shifts, in addition to in-distribution accuracy.

\section{Robustness in Vision vs. in NLP}

Despite the widely study of robustness in vision, 
the study of robustness in NLP cannot always directly borrow the ideas. 
We categorize the main differences with the three following 
points:

\paragraph{Continuous vs. Discrete in Search Space}
The most obvious characteristic is probably the discrete nature of the space of text. 
This particularly posed a challenge towards the adversarial attack and defense regime when the study in vision is transferred to NLP \cite{lei2019discrete, zhang2020adversarial}, 
in the sense that simple gradient-based adversarial attacks will not directly translate to meaningful attacks in the discrete text space, 
and multiple novel attack methods are proposed to fill the gap, as we will discuss in later sections. 

\paragraph{Perceptible to Human vs. Not}
On a related topic, one of the most impressive property of adversarial attack in vision is that 
small perturbation of the image data imperceptible to human are sufficient to deceive the model \cite{szegedy2013intriguing}, 
while this can hardly be true for NLP attacks. 
Instead of being imperceptible, the adversarial attacks in NLP typically are bounded by the fact that the meaning of the sentences are not altered (despite being perceptible).
On the other hand, 
there are ways to generate samples where the changes, 
although being perceptible, 
are often ignored by human brain due to some psychological prior on how a human processes the text \cite{anastasopoulos2019neural,wang2020word}. 

\paragraph{Support vs. Density Difference of the Data Distributions}
Another difference is more likely seen 
in the discussion of the domain adaptation of vision and NLP study. 
In vision study, although the images from training distribution and test distribution can be sufficiently different, the train and test distributions mostly share the same support
(the pixels are always sampled from a 0-255 integer space), although the density of these distributions can be very different (e.g., photos vs. sketches).
On the other hand, 
domain adaptation of NLP sometimes studies the regime where the supports of the data differ, e.g., the vocabularies can be significantly different in cross-lingual studies \cite{abad2020cross, zhang2020unsupervised}.

\paragraph{A Common Theme}
Despite the disparities between vision and NLP, 
the common theme of
pushing the model to generalize from $\mathcal{D}$ to $\mathcal{D}'$ preserves.
The practical difference between $\mathcal{D}$ and $\mathcal{D}'$ is more than often defined by the human's understanding of the data, and can differ in vision and NLP as humans perceive and process images and texts in subtly different ways, which creates both opportunities for learning and barriers for direct transfer.
Certain lines of research try to bridge the learning in the vision domain to the embedding space in the NLP domain, while other lines of research create more interpretable attacks in the discrete text space (see Table~\ref{tab:adversarial} for these two lines of work). How those two lines of research transfer to each other, or complement each other, is not fully explored and calls for additional research.

\begin{table*}[ht]
\centering
\small
    \begin{tabular}{c|c|c}
    \toprule
        Space & Perturbation level & Methods \\
       \midrule
        \multirow{5}{*}{Discrete} & Character-level & \makecell{HotFlip \cite{ebrahimi-etal-2018-hotflip}, DeepWordBug \cite{gao2018black}, \\Synthetic-Noise \cite{karpukhin2019training}} \\
        \cmidrule{2-3}
        & Word-level &  
        \makecell{GenAdv \cite{alzantot2018generating}, 
        PWWS \cite{ren-etal-2019-generating},\\
        SEM \cite{wang2019natural}, 
        {BERT}-{ATTACK} \cite{li-etal-2020-bert-attack}, \\
        TextFooler \cite{textfooler}, 
        SememePSO \cite{zang-etal-2020-word}}  \\
        \cmidrule{2-3}
        & Sentence-level & 
        \makecell{
        AdvSQuAD \cite{jia2017adversarial},
        SCPNs \cite{iyyer-etal-2018-adversarial}, \\ CAT-Gen \cite{wang-etal-2020-cat}, TAILOR \cite{ross2021tailor}}\\
        \cmidrule{2-3}
        & Mixed-types &  \makecell{
        CheckList \cite{ribeiro-etal-2020-beyond}, Polyjuice \cite{wu-etal-2021-polyjuice},\\ MAYA \cite{chen-etal-2021-multi}}\\
        \midrule
        Continuous & Embedding space & \makecell{AT \& VAT \cite{miyato_adversarial}, Natural-adversary \cite{natural_gan}, \\FreeLB \cite{freelb},  ALUM \cite{liu2020alum}}  \\
        \bottomrule
    \end{tabular}
\vspace{-0.05in}
    \caption{Perturbation types for identifying robustness failures and improving robustness in NLP.}
    \label{tab:adversarial}
\end{table*}

\begin{table*}[ht]
\centering
\small
    \begin{tabular}{c|c}
    \toprule
    Task & Robustness Benchmarks \\
    \midrule
    Natural Language Inference & \makecell{Stress-test \cite{naik2018stress}, HANS \cite{mccoy-etal-2019-right}, \\
    Counterfactual-NLI \cite{kaushik2019learning}, ANLI \cite{nie-etal-2020-adversarial}}\\
    \midrule
    Question Answering & \makecell{
    AdvSQuAD \cite{jia2017adversarial}, Adv-QA \cite{10.1162/tacl_a_00338}, \\
    Natural-Perturbed-QA \cite{khashabi-etal-2020-bang}, \\
    Natural-shift-QA \cite{pmlr-v119-miller20a}, SAM \cite{DBLP:conf/aaai/SchlegelNB21} }\\
    \midrule
    Paraphrase Identification & \makecell{PAWS \cite{zhang-etal-2019-paws}, PAWS-X \cite{yang-etal-2019-paws},\\ Modify-with-Shared-Words \cite{shi-huang-2020-robustness}} \\
    \midrule
    Co-reference & \makecell{WinoGender \cite{rudinger-EtAl:2018:N18}, WinoBias \cite{zhao-etal-2018-gender}} \\ 
    \midrule
    Named Entity Recognition & OntoRock \cite{lin-etal-2021-rockner}, SeqAttack \cite{simoncini-spanakis-2021-seqattack}\\
    \bottomrule
\end{tabular}
\vspace{-0.05in}
    \caption{A list of robustness benchmarks (challenging or adversarial datasets) and their corresponding tasks.}
    \label{tab:datasets}
\end{table*}

\section{Identify Robustness Failures}

As robustness gained increasing attention in NLP literature, various lines of work have proposed ways to identify robustness failures in NLP models. 
Existing works can be roughly categorized by \textit{how} the failures are identified, among which a large portion of work relies on human priors and error analyses over existing NLP models (Section~\ref{sec:human-prior}), and other lines of work adopt model-based approaches (Section~\ref{sec:model-driven-measure}).
The identified robustness failure patterns are usually organized into challenging/adversarial benchmark datasets to more accurately measure an NLP model's robustness. 
In Table~\ref{tab:adversarial}, we organize commonly used perturbation types for identifying models' robustness failures, and in Table~\ref{tab:datasets} we summarize common robustness benchmarks for each NLP task.

\subsection{Human Prior and Error Analyses Driven}
\label{sec:human-prior}
An increasing body of work has been conducted on understanding and measuring robustness in NLP models \cite{tu2020empirical,sagawa2020investigation,Geirhos_2020} across various NLP tasks, largely relying on human priors and error analyses.

\paragraph{Natural Language Inference}
\citet{naik2018stress} sampled misclassified examples and analyzed their potential sources of errors, which are then grouped into a typology of common reasons for error. Such error types then served as the bases to construct the \emph{stress test} set, to further evaluate whether NLI models have the ability to make real inferential decisions, or simply rely on sophisticated pattern matching.
\citet{gururangan2018annotation} found that current NLI models are likely to identify the label by relying only on the hypothesis, and 
\citet{poliak2018hypothesis} provided similar augments that using a hypothesis-only model  can outperform a set of strong baselines. 
\citet{kaushik2019learning} asked humans to generate counterfactual NLI examples, to better understand what features are causal and encourage models to learn those features.

\paragraph{Question Answering}
\citet{jia2017adversarial} proposed to generate adversarial QA examples
by concatenating an adversarial distracting sentence at the end of a paragraph. 
\citet{pmlr-v119-miller20a} built four new test sets for the Stanford Question Answering Dataset (SQuAD) and found most question-answering systems fail to generalize to this new data, calling for new evaluation metrics towards natural distribution shifts.

\paragraph{Machine Translation}
\citet{belinkov2018synthetic} found that character-based neural machine translation (NMT) models are brittle under noisy data, where noises (e.g., typos, misspellings, etc) are synthetically generated using possible lexical replacements. Data augmentation with artificially-introduced grammatical errors  \cite{anastasopoulos2019neural}
or with random synthetic noises  \cite{vaibhav2019improving,karpukhin2019training}
can make the system more robust to such spurious patterns. 
On the other hand, \citet{wang2020word} showed another approach by limiting the input space of the characters so that the models will be likely to perceive data typos and misspellings. 

\paragraph{Syntactic and Semantic Parsing}
Robust parsing has been studied in several existing works \cite{lee-etal-1995-robust,10.1017/S1351324902002887}. 
More recent work showed that neural semantic parsers are still not robust against lexical and stylistic variations, or meaning-preserving perturbations \cite{marzinotto-etal-2019-robust,huang-etal-2021-robustness}, and proposed ways to improve their robustness through data augmentation \cite{huang-etal-2021-robustness} and adversarial learning \cite{marzinotto-etal-2019-robust}.

\paragraph{Text Generation}
Existing work found that text generation models also suffer from robustness issues, e.g., text summarization models suffer from positional bias \cite{jung-etal-2019-earlier}, layout bias \cite{kryscinski-etal-2019-neural}, and a lack of faithfulness and factuality \cite{kryscinski-etal-2019-neural,maynez-etal-2020-faithfulness,chen-etal-2021-improving}; data-to-text models sometimes hallucinate texts that are not supported by the data \cite{parikh-etal-2020-totto,wang-etal-2020-towards}.
In addition, \citet{sellam-etal-2020-bleurt,BERTScore} pointed out the deficiency of existing automatic evaluation metrics and proposed new metrics to better align the generation quality with human judgements.

\paragraph{Connection with Dataset Biases}
The robustness failures can sometimes be attributed to dataset biases, i.e., biases introduced during dataset collection \cite{Fouhey18} or human annotation artifacts \cite{gururangan2018annotation,geva-etal-2019-modeling,rudinger-etal-2017-social}, which could affect how well a model trained from this dataset generalizes, and how accurately we estimate a model's performance.
For example, \citet{lewis-etal-2021-question} show there is a significant test-train data overlap in a set of open-domain question-answering benchmarks, and many QA models perform substantially worse on questions that cannot be memorized from training data.
In natural language inference, \citet{mccoy-etal-2019-right} show that commonly used crowdsourced datasets for training NLI models might make certain syntactic heuristics more easily adopted by statistical learners.
Further, \citet{pmlr-v119-bras20a} propose to use a lightweight adversarial filtering approach to filter dataset biases, which is approximated using each instance's predictability score.

\subsection{Model-based Identification}
\label{sec:model-driven-measure}

In addition to the human-prior and error-analysis driven approaches which are usually specific to each task, other lines of work identify robustness failures that are \textit{task-agnostic} like white-box text attack methods \cite{ebrahimi-etal-2018-hotflip,alzantot2018generating,textfooler}, and even \textit{input-agnostic} like universal adversarial triggers \cite{wallace-etal-2019-universal} and natural attack triggers \cite{song-etal-2021-universal}.

Another line of work proposes to learn an additional model to capture biases, e.g., in visual question answering, \citet{clark-etal-2019-dont} train a naive model to predict prototypical answers based on the question only irrespective of the context; \citet{he2019unlearn,utama-etal-2020-mind} propose to learn a biased model that only uses dataset-bias related features. This framework has also been used to capture unknown biases assuming that the lower capacity model learns to capture relatively shallow correlations during training \cite{clark2020learning}.
In addition, \citet{wang-culotta-2020-identifying} identify model shortcuts by training classifiers to better distinguish ``spurious'' correlations from ``genuine'' ones based on human annotated examples.

\paragraph{Model-in-the-loop vs. Human-in-the-loop}
Some work adopts human-in-the-loop to generate challenging examples, e.g., Counterfacutal-NLI \cite{kaushik2019learning} and Natural-Perturbed-QA \cite{khashabi-etal-2020-bang}.
Other work applies model-in-the-loop to increase the likelihood that the perturbed examples are challenging for state-of-the-art models, but it might also introduce biases towards the particular model used. For example, SWAG \cite{zellers-etal-2018-swag} was introduced that fooled most models at the time of publishing but was soon ``solved'' after BERT \cite{devlin-etal-2019-bert} was introduced. 
As a result, \citet{yuan-etal-2021-transferability} present a study over the transferability of adversarial examples, and Contrast Sets \cite{gardner-etal-2020-evaluating} intentionally avoid using model-in-the-loop. 
Further, more recent work adopts adversarial human-and-model-in-the-loop to create more difficult examples for benchmarking, e.g., Adv-QA \cite{10.1162/tacl_a_00338}, Adv-Quizbowl \cite{10.1162/tacl_a_00279}, ANLI \cite{nie-etal-2020-adversarial}, and Dynabench \cite{kiela-etal-2021-dynabench}.

\section{Improve Model Robustness}
Correspondingly, there are multiple lines of directions that try to improve robustness in NLP models. Depending on where and how the intervention is applied, those approaches can be categorized into the following categories: data-driven (Section~\ref{sec:data-driven}), model-based and training-scheme-based (Section~\ref{sec:model-training}), inductive-prior-based (Section~\ref{sec:inductive-prior}) and finally causal intervention (Section~\ref{sec:causal-intervention}).

\subsection{Data-driven Approaches}
\label{sec:data-driven}
Data augmentation recently gained a lot of interest, in improving performance in low-resourced language settings, few-shot learning, mitigating biases, and improving robustness in NLP models \cite{feng-etal-2021-survey,dhole2021nlaugmenter}.
Techniques like Mixup \cite{zhang2018mixup}, MixText \cite{chen2020mixtext}, CutOut \cite{devries2017improved},  AugMix \cite{hendrycks2020augmix}, HiddenCut \cite{chen2021hiddencut},  have been shown to substantially improve the robustness and the generalization of models.
Such mitigation strategies are operated at the data level, and often hard to be interpreted in terms of how and why mitigation works. 

Other lines of work deal with spans or regions associated within data points to prevent models from heavily relying on spurious patterns. 
To make NLP models more robust on sentiment analysis and NLI tasks, 
\citet{kaushik2019learning} proposed curating counterfactually augmented data via a human-in-the-loop process,
and showed that models trained on the combination of this augmented data and original data are less sensitive to spurious patterns. 
Differently, \citet{wang2021identifying} performed strategic data augmentation to perturb the set of   ``shortcuts'' that are automatically identified, and found that mitigating these leads to more robust models in multiple NLP tasks.  This line of mitigation strategies closely relates to how spurious correlations can be measured and identified, as many of the challenging or adversarial examples (Table~\ref{tab:adversarial}) can sometimes be used to augment the original model to improve its robustness, either in the discrete input space as additional training examples \cite{liu-etal-2019-inoculation,kaushik2019learning,anastasopoulos2019neural,vaibhav2019improving,khashabi-etal-2020-bang}, or in the embedding space \cite{freelb, natural_gan,miyato_adversarial,liu2020alum}.

\subsection{Model and Training-based Approaches}
\label{sec:model-training}
\paragraph{Pre-training}
Recent work has demonstrated pre-training as an effective way to improve NLP models' out-of-distribution robustness \cite{ood_robustness,tu2020empirical}, potentially due to its self-supervised objective and the use of large amounts of diverse pre-training data that encourages generalization from a small number of examples that counter the spurious correlations. \citet{tu2020empirical} showed a few other factors can also contribute to robust accuracy, including larger model size, more fine-tuning data, and longer fine-tuning.
A similar observation is made by \citet{taori2020measuring} in the vision domain, where the authors found training with larger and more diverse datasets offer better robustness consistently in multiple cases, compared to various robustness interventions proposed in the existing literature.

\paragraph{Training with a Better Use of Minority Examples}
Further, there are several works that propose to robustify the models via a better use of minority examples, e.g., examples that are under-represented in the training distribution, or examples that are harder to learn. For example, \citet{yaghoobzadeh-etal-2021-increasing} proposed to first fine-tune the model on the full data, and then on minority examples only.

In general, 
the training strategy with an emphasis on a subset of samples 
that are particularly hard for the model to learn 
is sometimes also referred to as group DRO \cite{Sagawa*2020Distributionally},
as an extension of vanilla distributional robust optimization (DRO) \cite{ben2013robust, duchi2021statistics}.
Extensions of DRO are mostly discussing the strategies on 
how to identify the samples considered as minority:
\citet{nam2020learning} trained two models in parallel, where the ``debiased'' model focuses on examples not learned by the ``biased'' model; 
\citet{NEURIPS2020_07fc15c9} used an adversary model to identify samples that are challenging to the main model;
\citet{pmlr-v139-liu21f} proposed to train the model a second time via up-weighting  examples that have high training losses during the first time.

\paragraph{When to Use Data-driven or Model-based Approaches?}
In many cases both the data and the model can contribute to a model's lack of robustness, hence data-driven and model-based approaches could be combined to further improve a model's robustness.
One interesting phenomenon observed by \cite{liu-etal-2019-inoculation} is to attribute models' robustness failures to blind spots in the training data, or the intrinsic learning ability of the model. The authors found that both patterns are possible: in some cases models can be inoculated via being exposed to a small amount of challenging data, similar to the data augmentation approaches mentioned in Section~\ref{sec:data-driven}; on the other hand, some challenging patterns remain difficult which connects to the larger question around generalizability to \textit{unseen} adversarial and counterfactual patterns \cite{huang-etal-2020-counterfactually,jha2020does,joshi2021investigation}, which is relatively under-explored but deserves much attention.

\subsection{Inductive-prior-based Approaches}
\label{sec:inductive-prior}
Another thread is to introduce inductive bias 
(i.e., to regularize the hypothesis space)
to force the model to discard some spurious features. 
This is closely connected to the human-prior-based identification approaches in Section~\ref{sec:human-prior} as those human-priors can often be used to re-formulate the training objective with additional regularizers.
To achieve this goal, 
one usually needs to first construct a side component 
to inform the main model about the misaligned features, 
and then to regularize the main model according to the side component. 
The construction of this side component 
usually relies on prior knowledge of what the misaligned features are.
Then, methods can be built accordingly to counter the features such as
label-associated keywords \cite{he2019unlearn},
label-associated text fragments \cite{MahabadiBH20}, 
and general easy-to-learn patterns of data \cite{nam2020learning}. 
Similarly, \citet{clark-etal-2019-dont,clark2020learning,utama-etal-2020-mind,utama-etal-2020-towards} propose to \textit{ensemble} with a model explicitly capturing bias, where the main model is trained together with this ``bias-only'' model such that the main model is discouraged from using biases.
More recent work \cite{xiong2021uncertainty} shows the ensemble-based approaches can be further improved via better calibrating the bias-only model.
Furthermore, additional regularizers have been introduced for robust fine-tuning over pre-trained models, e.g., mutual-information-based regularizers \cite{wang2021infobert} and smoothness-inducing adversarial regularization \cite{jiang-etal-2020-smart}.

In a broader scope, given
that one of the main challenges of domain adaptation is to 
counter the model's tendency in learning domain-specific spurious features \cite{GaninUAGLLML16}, 
some methods contributing to domain adaption
may have also progressed along the line of our interest, e.g.,  domain adversarial neural network \cite{GaninUAGLLML16}.
This line of work also inspires a family of methods 
forcing the model to learn auxiliary-annotation-invariant representations 
with a side component \cite{ghifary2016deep,WangMMX17,rozantsev2018beyond,motiian2017unified,li2018domain,WangWX19,Giorgos2020domain}. 

Despite the diverse concrete ideas introduced, the above is mainly training for small empirical loss across different domains or distributions in addition to forcing the model to be invariant to domain-specific spurious features. 
As an extension along this direction, 
invariant risk minimization (IRM) \cite{arjovsky2019invariant} introduces the idea of invariant predictors across multiple environments, which was later followed and discussed by a variety of extensions \cite{choe2020empirical,ahmed2020systematic,rosenfeld2020risks}. 
More recently, \citet{dranker2021irm} applied IRM in natural language inference and found that a more naturalistic characterization of the problem setup is needed.

\subsection{Causal Intervention}
\label{sec:causal-intervention}
Casual analyses have also been utilized to examine robustness. 
\citet{pmlr-v119-srivastava20a} leverage humans' common sense knowledge of causality to augment training examples with a potential unmeasured variable, and propose a DRO-based approach to encourage the model to be robust to distribution shifts over the unmeasured variables.
\citet{ananth-emnlp-2021} study the effect of secondary attributes, or confounders, and propose context-aware counterfactuals that take into account the impact of secondary attributes to improve models' robustness.
\citet{veitch2021counterfactual} propose to learn approximately counterfactual invariant predictors dependent on causal structures of the data, and show it can help mitigate spurious correlations in text classification.

\subsection{Connections between Mitigations}
Connecting these methods conceptually, 
we conjecture three different mainstream approaches:
one is to leverage the large amount of data by taking advantages of pre-trained models, another is to learn invariant representations or predictors across domains or environments,
while most of the rest build upon the prior on what the spurious patterns are and encourage the models to not rely on those patterns. Then the solutions are invented through countering model's learning of these patterns by either data augmentation, reweighting (the minorities), ensemble, inductive-prior design, and causal intervention. 
Interestingly, statistical work has shown that many of these mitigation methods are optimizing the same robust machine learning generalization error bound \cite{wang2021learning}. 

\section{Open Questions}
In addition to the challenges mentioned above, we list below a few open questions that call for additional research going forward.
\paragraph{Identifying Unknown Robustness Failures} 
Existing identification around robustness failures rely heavily on human priors and error analyses, which usually pre-define a small
or limited set of patterns that the model could
be vulnerable to. This requires extensive amount of expertise and efforts, and might still suffer from human or subjective biases in the end. How to 
proactively discover and identify models' unrobust
regions automatically and comprehensively remains challenging.

\paragraph{Interpreting and Mitigating Spurious Correlations} 
Interpretability matters for large NLP models, especially key to the robustness and spurious patterns. 
How can we develop ways to attribute or interpret these vulnerable portions of NLP models and communicate these robustness failures with designers, practitioners, and users? 
In addition, recent work \cite{wallace-etal-2019-allennlp,wang2021identifying,9521221} show interpretability methods can be utilized to better understand how a model makes its decision, which in turn can be used to uncover models' bias, diagnose errors,  and discover spurious correlations.

Furthermore, the mitigation of spurious correlations often suffers from the trade-off between removing shortcuts and sacrificing model performance \cite{NEURIPS2020_61d77652,zhang2019theoretically}.
Additionally, most existing mitigation strategies work in a pipeline fashion where defining and detecting spurious correlations are prerequisites, which might lead to error cascades in this process. How to design end-to-end frameworks for automatic mitigation deserves much attention. 

\paragraph{Unified Framework to Evaluate Robustness}
With a variety of potential spurious patterns in NLP models, it becomes increasingly challenging for developers and practitioners to quickly evaluate the robustness and quality of their models. This calls for more unified benchmarking efforts such as
CheckList \cite{ribeiro-etal-2020-beyond}, Reliability Testing \cite{tan-etal-2021-reliability}, Robustness Gym \cite{goel-etal-2021-robustness} and Dynabench \cite{kiela-etal-2021-dynabench}, to facilitate fast and easy evaluation of robustness. 

\paragraph{User Centered Measures and Mitigation}
Instead of passively detecting spurious correlations from a post-processing perspective, how to approach robustness from a user centric perspective needs further investigation. 
Based on the dual-process models of information processing, humans use two different processing styles \cite{evans2010intuition}. One is a quick and automatic style that relies on well-learned information and heuristic cues. The other is a qualitatively different style that is slower, more deliberative, and requires more reflective reasoning. Would these well-learned information and heuristic rules be leveraged to help design better human priors to measure and mitigate spurious correlations? If users or stakeholders are involved in this process, collecting a set of test cases where a system might perform well for the wrong reasons could help design sanity tests.

\paragraph{Connections between Human-like Linguistic Generalization and NLP Generalization}
\citet{linzen-2020-accelerate} argue NLP models should behave more like humans to achieve better generalization consistently.
It is interesting to note that how humans process information in NLP tasks exactly is still under exploration, and to what extent models should leverage human-knowledge is still a debatable topic.\footnote{\url{http://www.incompleteideas.net/IncIdeas/BitterLesson.html}} Nonetheless, if we can better understand and utilize the robustness properties in human perception, we can potentially advance models' robustness in a more meaningful way.

\section{Conclusion}
In this paper, we provided a unifying overview over robustness definitions, evaluations and  mitigation strategies in the NLP domain.
We also highlighted open challenges in this area to motivate future research, encouraging people to think deeply about more comprehensive benchmarks, transferability and validity of adversarial examples, unified framework to evaluate and improve robustness, user-centered measures and mitigation, and finally how to potentially achieve human-like linguistic generalization more meaningfully.

\section*{Acknowledgements} The authors would like to thank reviewers for their helpful insights and feedback. This work is  funded in part by a grant from Google.

\bibliographystyle{acl_natbib}
\bibliography{ref}
\end{document}